# Deep learning-based NLP Data Pipeline for EHR Scanned Document Information Extraction

Enshuo Hsu[1, 3, 4], Ioannis Malagaris[1], Yong-Fang Kuo[1], Rizwana Sultana[2], Kirk Roberts[3]

[1]Office of Biostatistics, [2]Division of Pulmonary, Critical Care and Sleep Medicine, Department of Internal Medicine, University of Texas Medical Branch, Galveston, Texas, USA. [3]School of Biomedical Informatics, University of Texas Health Science Center at Houston, Houston, Texas, USA. [4]Center for Outcomes Research, Houston Methodist, Houston, TX, USA.

## Abstract

Scanned documents in electronic health records (EHR) have been a challenge for decades, and are expected to stay in the foreseeable future. Current approaches for processing often include image preprocessing, optical character recognition (OCR), and text mining. However, there is limited work that evaluates the choice of image preprocessing methods, the selection of NLP models, and the role of document layout. The impact of each element remains unknown. We evaluated this method on a use case of two key indicators for sleep apnea, Apnea hypopnea index (AHI) and oxygen saturation ($SaO_2$) values, from scanned sleep study reports. Our data that included 955 manually annotated reports was secondarily utilized from a previous study in the University of Texas Medical Branch. We performed image preprocessing: gray-scaling followed by 1 iteration of dilating and erode, and 20% contrast increasing. The OCR was implemented with the Tesseract OCR engine. A total of seven Bag-of-Words models (Logistic Regression, Ridge Regression, Lasso Regression, Support Vector Machine, k-Nearest Neighbor, Naïve Bayes, and Random Forest) and three deep learning-based models (BiLSTM, BERT, and Clinical BERT) were evaluated. We also evaluated the combinations of image preprocessing methods (gray-scaling, dilate & erode, increased contrast by 20%, increased contrast by 60%), and two deep learning architectures (with and without structured input that provides document layout information). Our proposed method using Clinical BERT reached an AUROC of 0.9743 and document accuracy of 94.76% for AHI, and an AUROC of 0.9523, and document accuracy of 91.61% for $SaO_2$. We demonstrated the proper use of image preprocessing and document layout could be beneficial to scanned document processing.

## Introduction

Scanned documents in electronic health records (EHR) have been reported as a problem for almost a decade [1]. Generally these documents are the result of faxed medical records, which have been highly critized for decades [2] but they may also result from patients physically bringing medical records to the clinic. Regardless, solutions have focused on interoperability methods, from health information exchanges [3] to blockchain [4]. However, despite the availability of technical solutions, it seems clear that for the foreseeable future, scanned documents in the EHR will continue to play a prevalent part of our medical record ecosystem. It is thus critical to have informatics approaches that can process the information in scanned notes.

Common approaches to handling scanned documents include image preprocessing, optical character recognition (OCR), and text mining. Prior publications have reported promising results of adopting aspects of this workflow for real-world challenges [5]–[7]. However, there is limited work that evaluates the choice of image preprocessing methods, the selection of NLP models, and the role of document layout. The impact of each element and the interplay between them remains unexplored. Furthermore, as deep learning-based natural language processing (NLP) progresses, and new state-of-the-art language models based on Transformers [8] are introduced, scanned document information extraction studies have not kept pace with those advanced methods. Therefore, we propose a data pipeline for extracting important information from scanned reports in EHR. We evaluate this method on a use case of two key indicators for sleep apnea, Apnea hypopnea index (AHI) and oxygen saturation ($SaO_2$) values, from scanned sleep study reports. We evaluate five image preprocessing methods, seven machine learning-based bag-of-words models, and three deep learning-based sequence models. This study proposes the first data pipeline that adopted Transformer-based NLP models for scanned document information extraction, and to our knowledge is the first work that evaluates the impacts of image preprocessing methods, NLP model selection, and document layout utilization in scanned document processing for EHRs.

## Related Work

**Information extraction from scanned documents in EHRs.** Several reported scanned document processing studies focus on pathology and imaging reports that contain important clinical concepts and numeric values which are embedded in free-text narratives or non-standardized formats [5], [6], [9]. As a result of unintegrated EHR systems, those reports created by laboratories are often scanned and stored as images in Portable Document Format (PDF). Other sources of scanned documents include paper-based case report forms and outpatient referral forms which are created during hospital workflow that involves handwriting [7], [10]. Current approaches for processing scanned documents in EHR often involve two main steps: OCR and text mining. The OCR step extracts words from scanned images and converts them into machine-readable text, and the following text mining step extracts clinically relevant information from the text. A wide variety of OCR engines have been used in different studies including Adobe Acrobat Pro, FormScanner, and Tesseract. Text mining on the other hand tends to be homogenous. Most studies used rule-based algorithms either developed manually by domain experts in their organizations or from publicly available NLP tools. Besides the two main steps, assisting methods are sometimes applied to improve the overall performance. Image preprocessing that improves scanned image quality before the OCR has been applied to a few studies [7], [11]. Image segmentation isolates text components from the background. Gray-scaling reduces computation burden. Erosion regularizes the map of the text. Thresholding separates information from its background [12]. Those have been attempted to improve image quality in EHR scanned documents. However, limited work on evaluation has been done. On the other hand, post-OCR spelling correction is a more common method [6], [11], [13]. Based on some known commonly misrecognized letters (for example, "S" and "5", "i" and "1"), decision rules written in the regular expression are applied to the OCR outputs.

**Information extraction from scanned documents in non-clinical fields.** Handwriting recognition, scanned receipt information extraction, and automatic cheques processing is some applications of scanned document processing [12], [14]. In scanned receipt recognition, a recent study developed a processing pipeline that utilized deep learning approaches: the Connectionist Text Proposal Network (CTPN) for text detection, and the Attention-based Encoder-Decoder (AED) for text recognition [15]. The same study also evaluated image preprocessing using the receipt area against a threshold. In cheques recognition, a recent publication suggests that the use of a 2D Convolution Neural Network following image preprocessing using Otsu thresholding could achieve a 95.71% accuracy from a pool of sample cheques [16].

In summary, it is well-acknowledged that scanned documents still pose technical challenges for EHRs, as well as scientific challenges for how best to extract information from them. However, what is missing is an understanding of the interplay of how this information can be extracted, especially using modern machine learning-based NLP techniques. This is the gap this paper seeks to fill.

## Methods

### Data Source

We utilized a set of manually reviewed sleep study reports from an existing validation study in the University of Texas Medical Branch (UTMB) (IRB# 19-0189). In that prior study, the UTMB electronic health record (EHR) (Epic Systems) were queried for data from June 1, 2015, to May 31, 2018. A total of 3720 patients who had at least two outpatient visits to pulmonary clinics or primary care providers (PCP), were at least 18 years old, had at least one sleep disorder diagnosis code, and had BMI on record were included. The study randomly sampled 1200 patients (800 from pulmonary clinics and 400 from PCP) for a manual chart review, which was performed by a group of 4 sleep medicine specialists. Among the sampled patients, the AHI and $SaO_2$ values from 990 sleep study reports were found and recorded in a separate sheet. Some numeric values were rounded to integers during recording. Each report was only reviewed once by one of the 4 reviewers. Our study utilized the 990 reviewed reports and was approved by the institutional review board (IRB# 20-0266). This work was supported by a pilot grant of Biostatistics, Epidemiology and Research Design under UL1TR001439 from the National Center for Advancing Translation Sciences.

### Post-Annotation Processing

To utilize the secondary data for our purpose, we performed post-annotation processing. In the prior study, due to the lower requirement of precision, some decimal points were rounded to integers by reviewers. We recovered the original numeric values by looking up the scanned reports. We also excluded 35 reports without complete AHI and $SaO_2$ records. Our final material contains 2995 scanned images (in 955 unique reports) in Portable Document Format (PDF).

**Image Preprocessing**

We extracted images of scanned document pages from the PDF files followed by image preprocessing using the Open Source Computer Vision Library (OpenCV) [17]. We first convert images to gray-scale, then dilate and erode each character by 1 iteration of transformation [11] [18]. The dilate process results in the removal of small noise dots, while the erode process converts the image back to the original scale. Finally, we increased the contrast by 20% [19] (Figure 1).

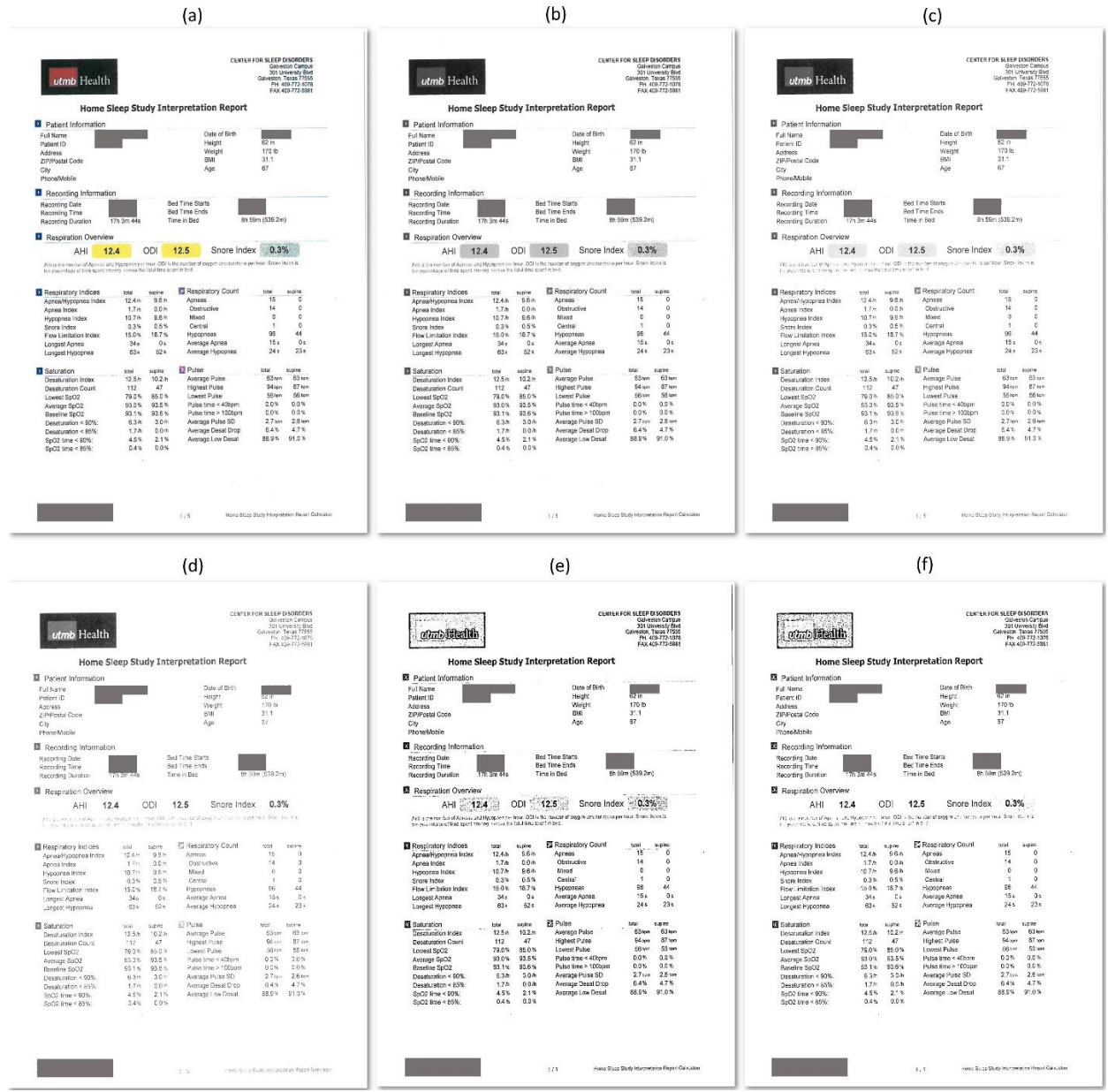

*Figure 1: Scanned document images after image preprocessing.* (a) The original scanned image. (b) the gray-scaled image. (c) the image with 20% increased contrast. (d) the image with 60% increased contrast. (e) the image with dilate & erode and 20% increased contrast. (f) the image with dilate & erode and 60% increased contrast.

**Optical Character Recognition (OCR)**

We applied the Tesseract OCR engine (version 4.0.0) [20] with a Python wrapper *pytesseract* [21] to locate and extract machine-readable text from the preprocessed images. The output for each image is a mapping of extracted words and positions in pixels. We performed a data quality visual inspection by programmatically drawing outlines of each word onto the original images using the positions with OpenCV (Figure 2).

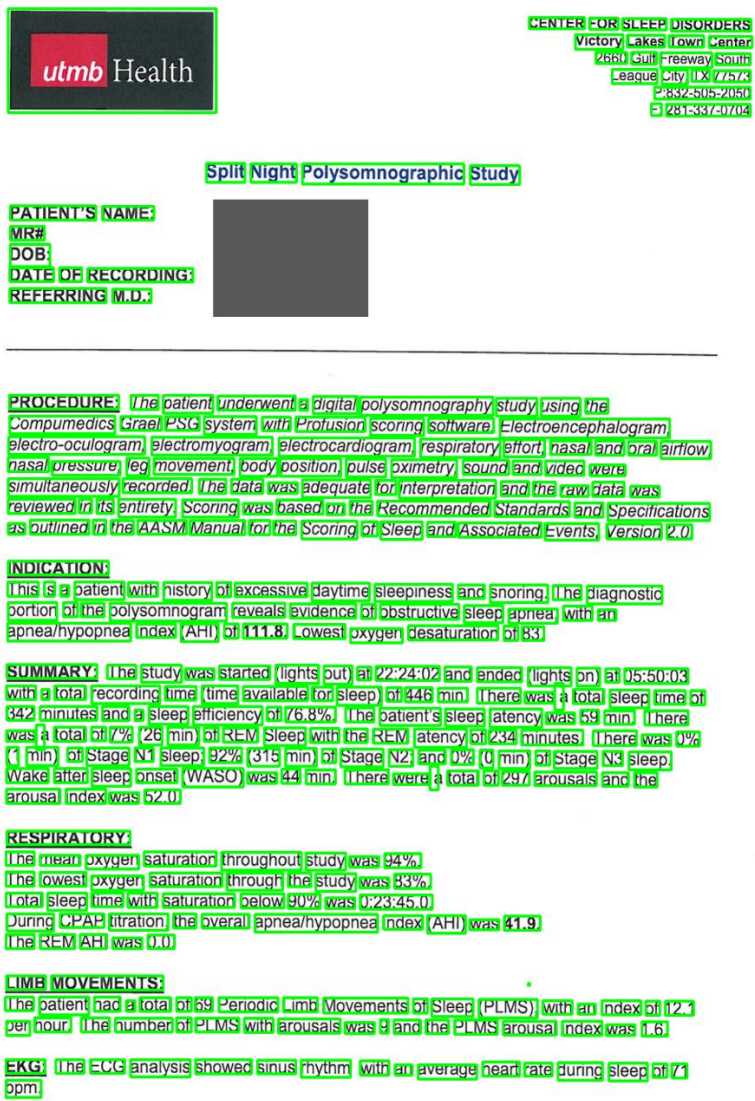

*Figure 2: Output of OCR for visual inspection*

**De-identification**

To ensure the confidentiality of patient information, we de-identified the output from OCR. We queried the EHR and created a lookup table that includes patient names and medical record numbers in each report. Then we used regular expressions to search among extracted words for those texts. The matched words were replaced by placeholders ("[PATNAME]", "[MRN]"). To exclude the date of birth and procedure dates, any output words with date formats ("XX/XX/XXXX") were replaced by placeholders ("[DATE]").

**Text Segmentation**

Each sleep study report on average (mean) has 3 pages with multiple paragraphs of free text. To reduce the amount of computation and ensure the efficiency of our NLP model, we programmatically identified candidate words for AHI and SaO$_2$ numeric values. We used a regular expression to search for any words that match "[0-9.,%]+". For each numeric value, we captured a segment of 10 words on each side of the candidate (21 words total). Examples are shown in Table 1.

*Table 1: Analytical data for text classification.*

| left | top | width | height | page | Numeric value | Segment | label |
|---|---|---|---|---|---|---|---|
| 735 | 388 | 61 | 26 | 1 | 26.0 | hypopneas, 120 met the AASM Version 2 scoring rule, while **26** met the Medicare scoring rule. The total APNEA/HYPOPNEA INDEX (AHI) | Other |
| 1048 | 385 | 111 | 50 | 1 | 19.5 | the Medicare scoring rule. The total APNEA/HYPOPNEA INDEX (AHI) was **19.5.** The patient also had 0 respiratory event related arousals (RERA) | AHI |
| 232 | 456 | 150 | 25 | 1 | 120.0 | and an apnea index of 1.3. There were 129 hypopneas, **120** met the AASM Version 2 scoring rule, while 26 met | Other |

The column "left" and "top" are the coordinate of pixels for the top-left corner of the word regions.
The column "width" and "height" are the width and height in pixels of the word regions.
The column "page" indicates from which page of the document the numeric value was extracted.
The column "Numeric value" is the float point representation of the numeric value.
The column "Segment" holds the free text segment of 21 words.
The column "Label" was derived from manual chart review and was used as the label for the supervised learning classifiers.

**Text Classification**

At this point, the information extraction problem can now be cast into a three-way classification task: whether the candidate numeric value is an AHI value, a SaO$_2$ value, or neither. Each instance has a set of position indicators obtained from OCR, the page number from which the numeric value was extracted, a floating-point representation of the numeric value, a segment composed of 21 words, and a label indicating that the numeric value was "AHI", "SaO$_2$" or "Other". Our human review did not include information on positions where the AHI and SaO$_2$ values were found. We assigned labels by matching the recorded AHI and SaO$_2$ numbers to each of the numeric values in the document. Therefore, as a limitation, we could not rule out false positives if some other numeric values in the same report happened to be the same number as the AHI or SaO$_2$, though we suspect this to be quite rare.

In our main experiment, we constructed and trained two types of NLP models: bag-of-words models and deep learning-based sequence models. The methodology flowchart is shown in Figure 4.

**Bag-of-Words Models**

A bag-of-words model that only considers term frequency is the traditional approach for text classification. We preprocessed the segments by removing English stopwords and converting all text to lowercase with Natural Language Toolkit [22]. Term frequency-inverse document frequency (tf-idf) was calculated for the top 400 words with the highest term frequencies in the training set, followed by a vector normalization [23] [24]. The features for the classifiers were:
1. four position indicators obtained from the OCR
2. page number
3. float point representation of the numeric value
4. tf-idf of the top 400 terms

We examined well-established machine learning classifiers including Logistic Regression, Ridge Regression, Lasso Regression, Support Vector Machine, k-Nearest Neighbor, Naïve Bayes, and Random Forest. All models were implemented in *Scikit-learn* [25].

**Deep Learning-based Sequence Models**

In recent years, deep learning adoption in clinical NLP publications has grown substantially [26]. To evaluate the efficacy of these models for our task, we examined Bidirectional Long short-term memory (BiLSTM), Bidirectional Encoder Representations from Transformers (BERT) [8], and the secondarily pre-trained BERT using EHR data (Clinical BERT) [27]. All deep learning models are evaluated as a component of a parent neural network architecture shown in Figure 3. The network includes an input branch for the structured features (position indicators, page number, numeric value in float). The inputs are batch-normalized [28], and passes to two layers of a feed-forward neural network (FFNN) with 100 neurons and a 20% dropout rate in each layer. The network also includes an input branch for segments (sequences). The sequence features input with a maximum sequence length of 32 tokens, is encoded, processed (with BiLSTM, BERT, or Clinical BERT), flattened, and passed to an FFNN. The structured and sequence input branches are concatenated and fully connected to the classifier layers which includes a FFNN with 200 neurons and a 20% dropout rate, followed by the output layer with a sigmoid activation function. The outputs are multinomial with probabilities for the 3 categories: "AHI", "SaO$_2$" and "Other". All deep learning models are constructed with TensorFlow [29] and Keras [30].

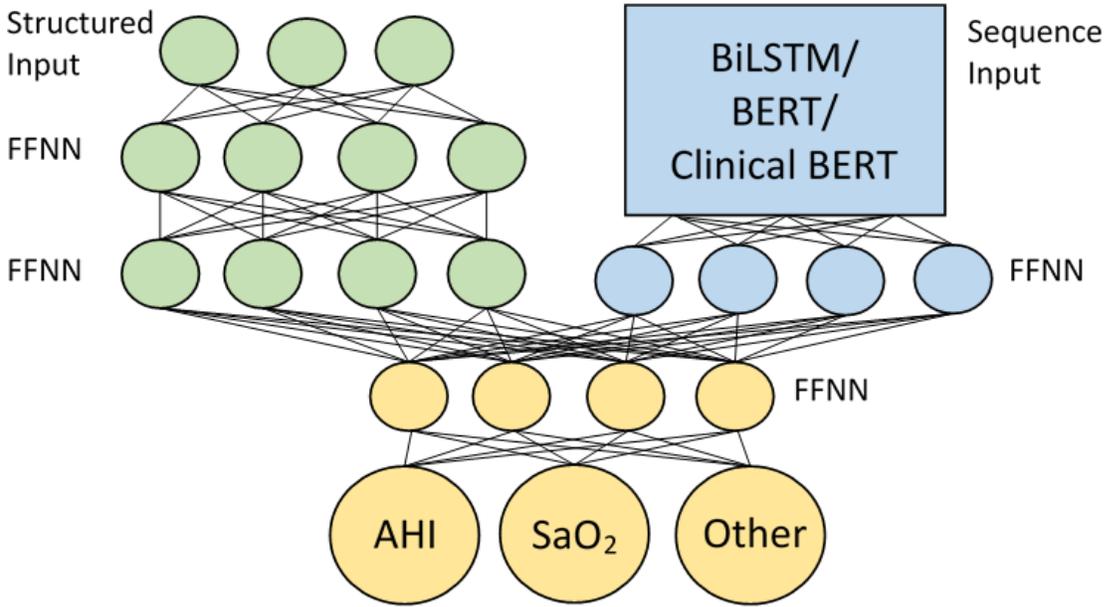

*Figure 3: Parent neural network architecture.* The structured input branch (green) takes in position indicators, page number, and numeric value. The sequence input branch (blue) takes in encoded segments, processed by specific deep learning architectures, and flattened to remove time steps. The classifier layers (yellow) connect the structured input branch (green) and sequence input branch (blue) and make predictions.

For BiLSTM, we used the word2vec [31] embedding which was implemented with the *Gensim* library [32] and was pre-trained on the training set using Continuous bag-of-word (CBOW) [33] via unsupervised learning. We applied an embedding dimension of 100. We then input the embedded word vectors to the model through 2 layers of BiLSTM where the second layer's last hidden state is fed to the classifier layers.

We implemented the BERT model by combining the uncased BERT-base model [34] in the *transformers* library [35] with TensorFlow. The segments were first tokenized and embedded with WordPiece embedding [36] before input into the BERT model. We flattened the outputs from BERT (a vector of 768 dimensions for each of the 32 input tokens) and passed them to an FFNN, followed by the classifier layers.

Clinical BERT was implemented similarly to BERT. We loaded the Bio_Clinical BERT model [37] in the *transformers* library through PyTorch [38], and converted it to a TensorFlow model using the *Transformers* library.

**Model Training and Evaluation**

To examine the NLP models, we split the reports into a 70% ($N$=669) development set and a 30% test set ($N$=286). For the bag-of-words models, we performed 5-fold cross-validation using the development set to search for an optimal parameter set that maximizes the validation accuracy. We then re-trained each model with the entire development set given the optimal parameters. For the deep learning-based sequence models, due to the high demand for computation, we further split the 70% development set with a 6:1 ratio into a training set ($N$=574) and a validation set ($N$=95). We saved checkpoints after each epoch and used the validation set to select the best checkpoint as our final model, based on cross-entropy loss. The BiLSTM model was trained using a batch size of 64, with Adam optimization with a learning rate of 2e-4 for 100 epochs. The BERT and Clinical BERT models were fine-tuned using a batch size of 64, with Adam optimization with a learning rate of 2e-6 for 100 epochs.

After training, the final models were evaluated with the test set. We evaluated at the segment level using recall, precision, and the area under the receiver operating characteristic curve (AUROC) for AHI and $SaO_2$. To better assess our final goal for information extraction, we also evaluated it at the document level. The numeric value in a document with the highest probability for AHI (or $SaO_2$) was selected to represent the document. We define document accuracy as:

$$\text{document accuracy} = \frac{\#\ of\ documents\ correctly\ extracted}{\#\ of\ documents\ in\ test\ set}$$

We performed DeLong's test [39] for comparing AUROC, and the Chi-square test for comparing document accuracy among the models. Family-wise error rates were adjusted using the Bonferroni procedure.

**Training Set Size Effect Analysis**

To assess the effect of a training set size on the model performance, our second experiment focused on subsets of the training set. We independently sampled from the training set (*N*=574) and built subsets of 10 reports, 25 reports, 50 reports, and 100 reports. We used each of the subsets to train the BiLSTM, BERT, and Clinical BERT models and used the validation set (*N*=95) to select final models based on the cross-entropy loss. The final models were evaluated with the test set (*N*=286).

**Stand-alone Validation Analyses**

To explore the impact of image preprocessing on the final performance, we examined 6 different image preprocessing methods: 1) gray-scale (baseline), 2) gray-scale + dilate/ erode, 3) gray-scale + increase contrast by 20%, 4) gray-scale + increase contrast by 60%, 5) gray-scale + dilate/ erode + increase contrast by 20% (our proposed method), 6) gray-scale + dilate/ erode + increase contrast by 60%. Figure 1 shows the output images on a scanned report. For each preprocessing method, OCR was performed followed by a sequence model with Clinical BERT to evaluate performances.

Our proposed sequence models involved both structured and sequence features. To evaluate the contribution of structured features, we examined an architecture with only the sequence input branch (excluding the structured input branch). Clinical BERT was used followed by the classifier layers.

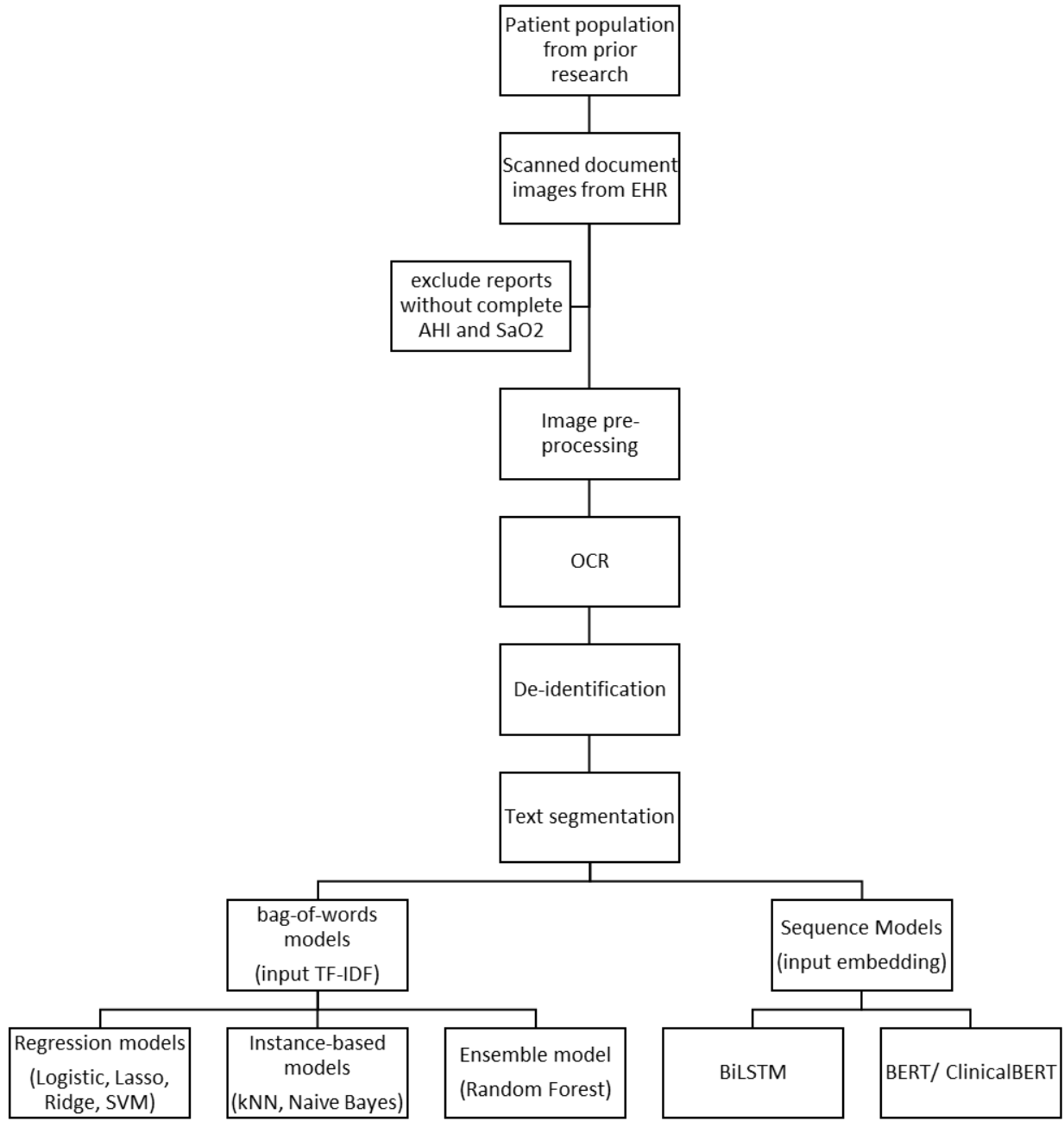

*Figure 4: Data pipeline flowchart*

## Results

The sleep study reports were generated by different laboratories in various structures and layouts (Figure 5). From visual inspection of the original documents, most findings were reported in narratives in the printed text. Besides, the reports also contained images (hospital logos, figures, and plots), tables, and handwritings. There were physician signatures and handwritten notes on the edges in some reports. Several reports from a certain laboratory have a similar structure of paragraphs and sentences, indicating a likelihood of templates being used.

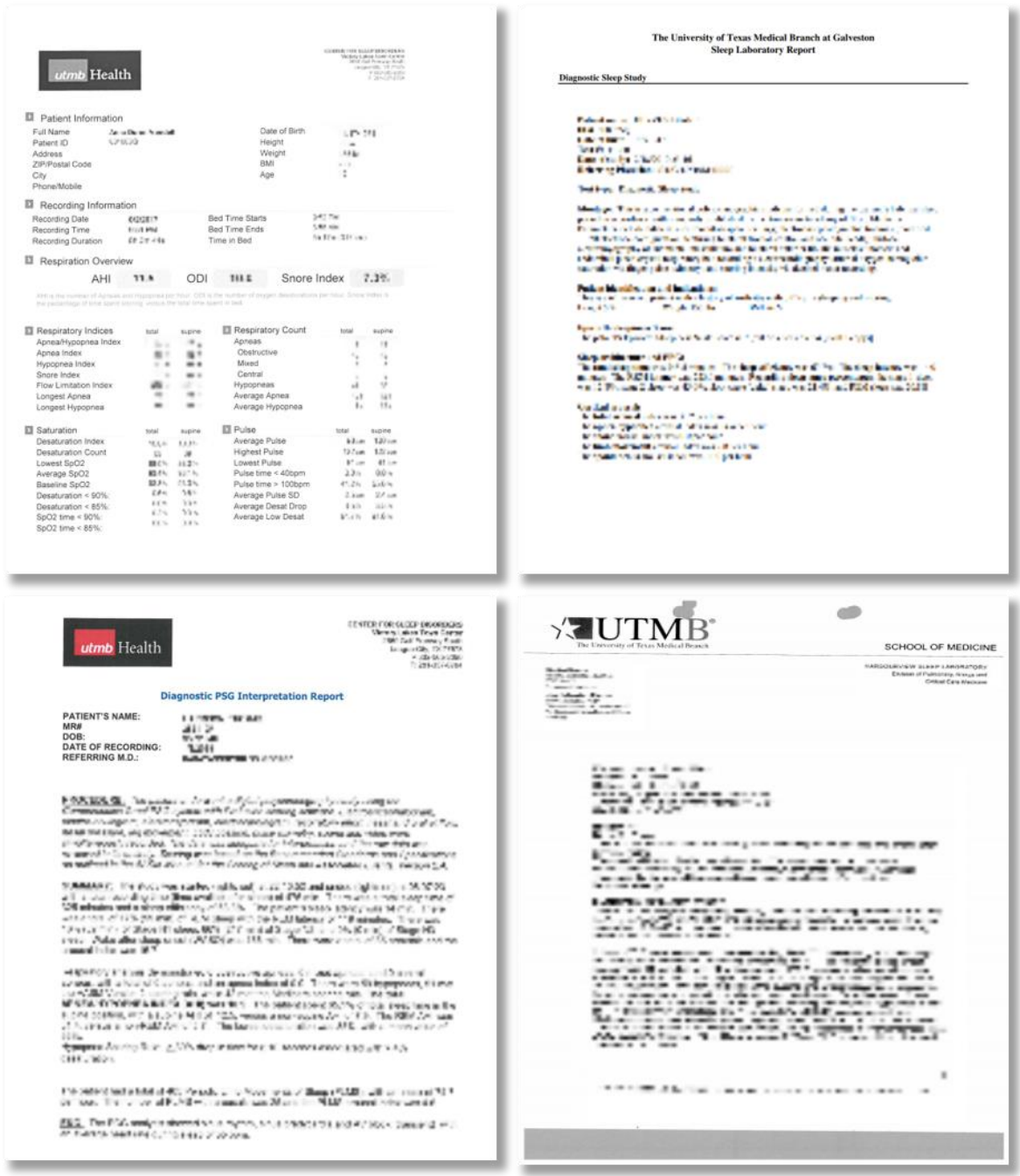

*Figure 5: Collection of scanned sleep study reports.* *The images have been intentionally blurred, their purpose here is to provide a sense of the overall structure and consistency (and lack thereof) between scanned documents.*

In the post-OCR inspection, we found most printed text, either in paragraphs or in tables, could be located. However, figures and handwriting added complexity to the algorithm. We noticed that in some reports, parts of images were considered text. Also, as reported in previous studies [11] [1], we noticed some misspelled words in the outputs. For example, the letter "I" was recognized as "!" or ")".

The scanned reports had a median of 2 pages (Q1-Q3 = [2, 4], range = [1, 28]). The reports included a median of 44 numeric values (Q1-Q3 = [37, 96]). About 52% of the reports had multiple numeric values that were labeled as AHI (median = 2, Q1-Q3 = [1, 2]) (Table 2).

*Table 2: Summary of data and labels*

|  | PDF documents | | OCR outputs | | | |
|---|---|---|---|---|---|---|
|  | **Reports** | **Pages** | **Numeric values** | **Instances of AHI** | **Instances of $SaO_2$** | **Instances of Other** |
| **Entire data set** | 955 | 2975 | 80148 | 1877 | 1668 | 76603 |
| **development set** | 669 | 2031 | 56839 | 1323 | 1146 | 54370 |
| **Test set** | 286 | 957 | 27076 | 581 | 552 | 25943 |

In our first experiment, the deep learning-based sequence models in general performed better than the bag-of-words models. For extraction of AHI, most bag-of-words models had high precisions at the segment level (from 0.4367 to 0.9865) that were close to sequence models (from 0.8803 to 0.9843). But sequence models had much higher recalls (from 0.6454 to 0.7470) compared to bag-of-words models (from 0.4802 to 0.6713). BERT and Clinical BERT showed the highest AUROC of 0.9705 and 0.9743, respectively. At the document level, the best word-bag models, kNN and Random Forest had around 93.5% accuracy while BERT and Clinical BERT reached 94% to 95% accuracy (Table 3, Figure 6).

For $SaO_2$ extraction, we found similar patterns as AHI. Sequence models had a much higher recall (from 0.6739 to 0.7319) at the segment level. Clinical BERT achieved the highest AUROC of 0.9523. At the document level, sequence models had higher accuracy of 91.6% while bag-of-words models' accuracy ranged from 51.8% to 89.5% (Table 3).

Comparing among sequence models, for AHI extraction, Clinical BERT had significantly higher AUROC than BiLSTM (p=.0008). For $SaO_2$ extraction, Clinical BERT achieved the highest AUROC and was significantly higher than BERT (p=0.0029) and BiLSTM (p<.0001). We did not see any significant document accuracy differences given the limited sample size in the test set (Table 4).

Summarizing the first experiment, our data pipeline with Clinical BERT as the NLP model could achieve the best performance for AHI extraction (AUROC of 0.9743, 94.76% document accuracy) and $SaO_2$ extraction (AUROC of 0.9523, 91.61% document accuracy).

*Table 3: Evaluation of different classifiers*

| | Classifier | Segment-level | | | Document-level |
|---|---|---|---|---|---|
| | | Recall | Precision | AUROC | Accuracy (%) |
| **AHI** | | | | | |
| Bag-of-words models | LR | 0.4819 | 0.8383 | 0.9093 (0.8932 - 0.9254) | 87.41 (83.57 - 91.26) |
| | LASSO (L1) | 0.4819 | 0.8889 | 0.9169 (0.9014 - 0.9325) | 89.16 (85.56 - 92.76) |
| | Ridge (L2) | 0.4802 | 0.8429 | 0.9176 (0.9021 - 0.9331) | 87.41 (83.57 - 91.26) |
| | SVM | 0.6093 | 0.9752 | 0.9050 (0.8886 - 0.9215) | 93.01 (90.05 - 95.96) |
| | kNN | 0.6713 | 0.8534 | 0.8644 (0.8454 - 0.8834) | 93.57 (90.36 - 96.78) |
| | NaiveBayes | 0.5577 | 0.4367 | 0.9179 (0.9024 - 0.9334) | 75.87 (70.92 - 80.83) |
| | Random Forest | 0.6299 | 0.9865 | 0.9476 (0.9350 - 0.9603) | 93.71 (90.89 - 96.52) |
| Sequence models | BiLSTM | 0.6454 | 0.9843 | 0.9637 (0.9530 - 0.9743) | 94.06 (91.32 - 96.80) |
| | BERT | 0.747 | 0.8803 | 0.9705 (0.9609 - 0.9802) | **95.10 (92.60 - 97.61)** |
| | Clinical BERT | 0.7315 | 0.914 | **0.9743 (0.9652 - 0.9833)** | 94.76 (92.17 - 97.34) |
| **SaO$_2$** | | | | | |
| Bag-of-words models | LR | 0.567 | 0.4914 | 0.9153 (0.8996 - 0.9309) | 82.87 (78.50 - 87.23) |
| | LASSO (L1) | 0.538 | 0.5103 | 0.9151 (0.8994 - 0.9308) | 84.62 (80.43 - 88.80) |
| | Ridge (L2) | 0.5543 | 0.4904 | 0.9143 (0.8985 - 0.9300) | 83.22 (78.89 - 87.55) |
| | SVM | 0.6105 | 0.9133 | 0.8860 (0.8683 - 0.9038) | 87.76 (83.96 - 91.56) |
| | kNN | 0.587 | 0.8663 | 0.8429 (0.8228 - 0.8629) | 87.86 (83.84 - 91.88) |
| | NaiveBayes | 0.6322 | 0.2705 | 0.9082 (0.8919 - 0.9244) | 51.75 (45.96 - 57.54) |
| | Random Forest | 0.6087 | 0.9307 | 0.9264 (0.9117 - 0.9412) | 89.51 (85.96 - 93.06) |
| Sequence models | BiLSTM | 0.6739 | 0.9051 | 0.9274 (0.9127 - 0.9420) | **91.61 (88.40 - 94.82)** |
| | BERT | 0.7319 | 0.8651 | 0.9358 (0.9219 - 0.9497) | **91.61 (88.40 - 94.82)** |
| | Clinical BERT | 0.683 | 0.8871 | **0.9523 (0.9402 - 0.9644)** | **91.61 (88.40 - 94.82)** |

Logistic Regression does not apply penalty; Lasso regression has L1 penalty ($\lambda = 0.01$); Ridge has L2 penalty ($\lambda = 0.01$). Support Vector Machine uses a polynomial kernel. kNN uses k = 3. NaiveBayes classifier uses alpha = 0.5. BiLSTM uses Word2Vec model for embedding pre-trained on the training set with CBOW, input vector of 100 dimensions. BERT and Clinical BERT are fine-tuned for 100 epochs with sequence length 32, and batch size 64.

*Table 4: Comparing Clinical BERT with BERT, BiLSTM, and Random Forest*

| Adjusted p-value | Clinical BERT vs. BERT | | Clinical BERT vs. BiLSTM | | Clinical BERT vs. Random Forest | |
|---|---|---|---|---|---|---|
| | AHI | SaO$_2$ | AHI | SaO$_2$ | AHI | SaO$_2$ |
| AUROC | 0.4528 | **0.0029** | **0.0008** | **<.0001** | **<.0001** | **<.0001** |
| Document accuracy | 1.0000 | 1.0000 | 1.0000 | 1.0000 | 1.0000 | 1.0000 |

AUROC was pair-wisely compared with DeLong's test. Document accuracy was pair-wisely compared with the Chi-square test. All p-values are corrected with the Bonferroni procedure.

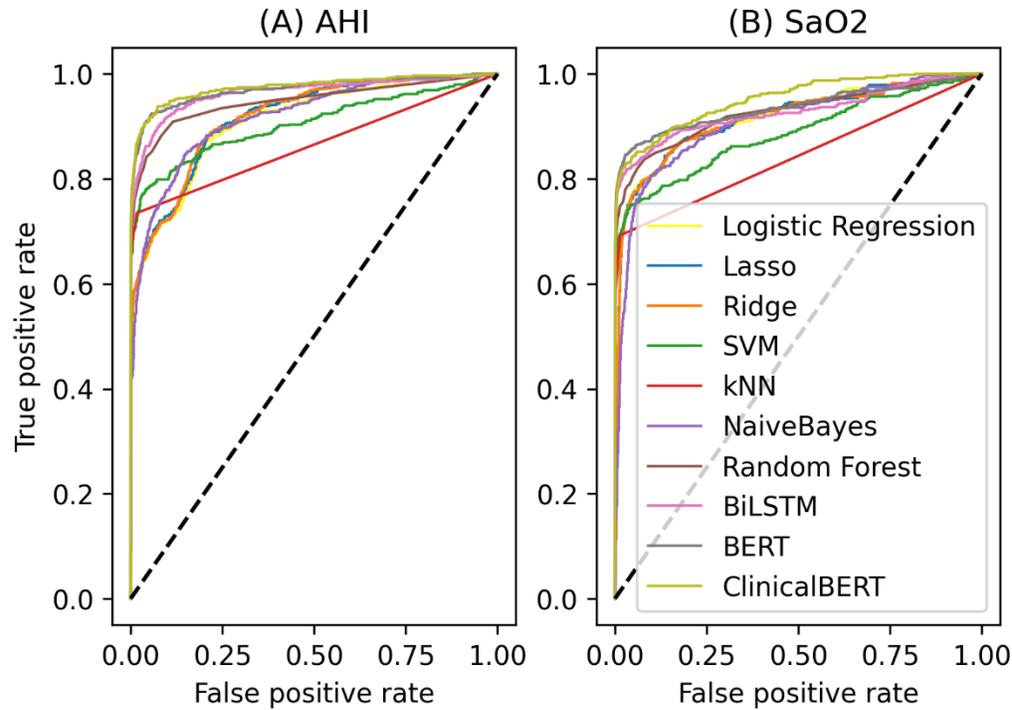

*Figure 6: ROC curve for each classifier*

In our second experiment, we examined the effect of different training set sizes on model performance. The purpose is to reflect real-world conditions where a large amount of training data is often unavailable. For AHI extraction, Clinical BERT reached an AUROC of 0.8612 and a document accuracy of 75% training on only 25 reports. While with the same sample size, BERT had AUROC of 0.8501 and a document accuracy of 73%. BiLSTM had an AUROC of 0.7954 and a document accuracy of 29%. When the sample size was 50 reports, the 3 deep learning-based sequence models had similar performance around 0.9 of AUROC and 90% document accuracy. For $SaO_2$ extraction, the Clinical BERT and BERT had similar performances. Trained on 25 reports, they achieved an AUROC of 0.8333 and 0.8458, and a document accuracy of 81% and 83% respectively, while BiLSTM had an AUROC of 0.7279 and a document accuracy of 18%. With 50 reports as the training set, all 3 models achieved AUROC of 0.88 and 85% document accuracy (Figure 7). Summarizing the second experiment, Clinical BERT could perform better with less training data. However, with at least 50 reports in the training set, all 3 models could perform similarly well.

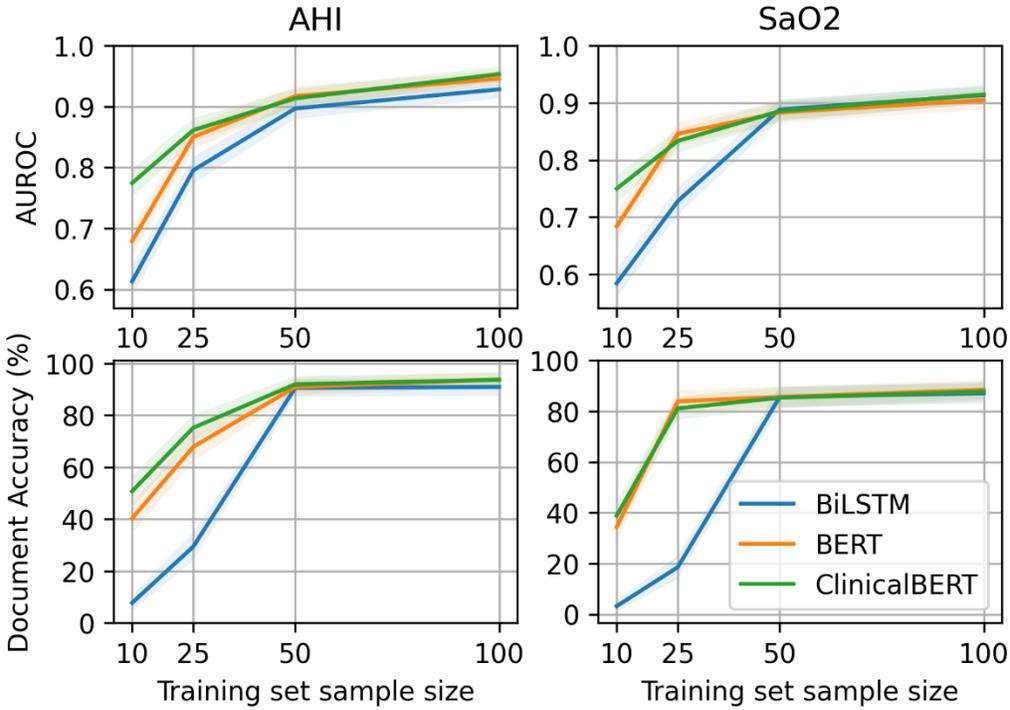

*Figure 7: Evaluation of effects of training set size*

As a stand-alone validation analysis, we evaluated different image preprocessing methods followed by Clinical BERT. The results showed 1 iteration of dilating and eroding with an increased contrast of 20% resulted in the best performance for AHI extraction and second best performance for SaO$_2$ in AUROC (Table 5).

*Table 5: Comparison of different image preprocessing methods*

|  | Image preprocessing | Segment-level | | | Document-level |
|---|---|---|---|---|---|
|  |  | Recall | Precision | AUROC | Accuracy (%) |
| **AHI** | Gray scale | 0.7187 | 0.9249 | 0.9699 (0.9601 - 0.9796) | 95.45 (93.04 - 97.87) |
|  | Gray scale + dilate & erode | 0.6961 | 0.9687 | 0.9679 (0.9573 - 0.9784) | 94.41 (91.74 - 97.07) |
|  | Gray scale + contrast 20% | 0.7126 | 0.9324 | 0.9705 (0.9609 - 0.9802) | 94.06 (91.32 - 96.80) |
|  | Gray scale + contrast 60% | 0.7268 | 0.9216 | 0.9692 (0.9593 - 0.9790) | 95.45 (93.04 - 97.87) |
|  | Gray scale + dilate & erode + contrast 20% | 0.7315 | 0.914 | **0.9743 (0.9652 - 0.9833)** | 94.76 (92.17 - 97.34) |
|  | Gray scale + dilate & erode + contrast 60% | 0.7268 | 0.9216 | 0.9692 (0.9593 - 0.9790) | **95.80 (93.48 - 98.13)** |
| SaO$_2$ | Gray scale | 0.7258 | 0.8617 | 0.9334 (0.9193 - 0.9474) | 91.61 (88.40 - 94.82) |
| | Gray scale + dilate & erode | 0.7427 | 0.8819 | **0.9620 (0.9506 - 0.9734)** | 90.21 (86.77 - 93.65) |
| | Gray scale + contrast 20% | 0.6957 | 0.8889 | 0.9431 (0.9300 - 0.9563) | 91.26 (87.99 - 94.53) |
| | Gray scale + contrast 60% | 0.6863 | 0.8671 | 0.9495 (0.9370 - 0.9619) | 91.61 (88.40 - 94.82) |
| | Gray scale + dilate & erode + contrast 20% | 0.683 | 0.8871 | 0.9523 (0.9402 - 0.9644) | 91.61 (88.40 - 94.82) |
| | Gray scale + dilate & erode + contrast 60% | 0.6863 | 0.8671 | 0.9495 (0.9370 - 0.9619) | **91.96 (88.81 - 95.11)** |

We also evaluated different sequence model architectures: with structured inputs and without structured inputs using Clinical BERT. The results showed adding structured input would improve AUROC by 0.0043 for AHI and 0.0107 for SaO$_2$ and document accuracy of 0.35% for AHI and 0.7% for SaO$_2$ (Table 6).

*Table 6: Comparison of different sequence model architectures*

| | Model Architecture | Segment-level | | | Document-level |
|---|---|---|---|---|---|
| | | Recall | Precision | AUROC | Accuracy (%) |
| AHI | sequence input | 0.7522 | 0.8723 | 0.9703 (0.9606 - 0.9800) | 94.41 (91.74 - 97.07) |
| | sequence input + structured input | 0.7315 | 0.914 | **0.9743 (0.9652 - 0.9833)** | **94.76 (92.17 - 97.34)** |
| SaO$_2$ | sequence input | 0.692 | 0.8761 | 0.9430 (0.9298 - 0.9561) | 90.91 (87.58 - 94.24) |
| | sequence input + structured input | 0.683 | 0.8871 | **0.9523 (0.9402 - 0.9644)** | **91.61 (88.40 - 94.82)** |

## Discussion

Our proposed data pipeline with appropriate image preprocessing (gray-scaling, dilate & erode, and increased contrast by 20%) and the Clinical BERT sequence model with structured input features showed positive performance for extracting laboratory result values from scanned documents (AUROC of 0.9743; document accuracy of 94.76% for AHI, and AUROC of 0.9523; document accuracy of 91.61% for SaO$_2$). Given the fact that all our annotated data was secondarily imported from a previous study for different purposes, we demonstrated a successful example of how scanned document processing can be accomplished in a real-world scenario. Manual annotation is time-consuming and expensive. In today's hospital systems where scanned document processing in EHRs is eagerly needed, it is often a challenge that compromises research feasibility. Our study design minimized the need for manual annotation. We utilized existing labeled reports, created annotated segments with automatic value-matching programs, trained NLP models, and proposed a data pipeline that extracted the needed variables with high performance. Our sample size experiment further showed that the BERT-based sequence models could achieve 90% document accuracy for AHI trained only on 50 annotated reports. While training on the entire 574 training set resulted in 95% document accuracy. This implied that depending on the nature of each downstream task, there is flexibility for trading a small degree of performance for a significant drop in annotation costs.

To analyze the difficulty of the NLP piece of the pipeline, we evaluated 7 bag-of-words NLP models and 3 deep learning-based sequence NLP models. Our evaluation showed that Clinical BERT could achieve the best AUROC and document accuracy. This finding is consistent with a previous study that utilized Clinical BERT for scanned document classification [11]. The authors reported an accuracy of 97.3% for classifying clinically relevant documents versus not clinically relevant documents, while the Random Forest classifier achieved only 95.8%. The authors did not evaluate other deep learning models. Our evaluation covered a wider range of machine learning and deep learning methods. We reported that BiLSTM, BERT, and Clinical BERT performed better than bag-of-words models. Among bag-of-words models, Random Forest performed the best. Our study is one of the first that evaluated deep learning-based NLP models for scanned document processing, though we also demonstrate that other aspects of the scanned document pipeline are important as well.

Though the open-source OCR library Tesseract has proven effective, appropriate image preprocessing is needed for realistic input [40]. Few publications have focused on evaluating different image preprocessing methods for scanned medical documents. An earlier study that focused on evaluating OCR engines for scanned medical documents applied image preprocessing methods for overlapping-lines removal [1]. In our documents, since the majority of the text was printed instead of hand-written, overlapping lines were not a significant issue. We referenced a recent study that utilized gray-scaling, erosion, and increasing contrast to improve image quality [11]. Gray-scaling has been reported to improve Tesseract OCR performance. Dilation, erosion, and contrast are simple transformations that have a uniform effect when applied to images with different scanned quality, compared to complicated thresholding transformations. Thus, we chose these methods for data pipeline development. We believe our work is a practical starting point. To better validate the selection of image preprocessing methods, scanned documents with annotated text are needed. Some related studies included a post-OCR text processing that involved spelling check [11][6]. We omitted this step considering the inconsistent abbreviations in our documents.

The presence of layout and word position is a unique feature in scanned text compared to digitally stored text in EHR. In the scanned sleep study reports, there was often an organization logo on the top left, a title line in the middle, and a facility name and contact information on the top right. Important findings were often placed in obvious positions, for example, the upper part of the first page. We noticed some laboratories have developed their internal reporting templates. This also contributes to the consistency of word positioning. Therefore, when developing the data pipeline, we utilized the word positions and page numbers as additional information. Our validation analysis found that adding structured inputs including word position in pixels, page number, and numeric value could boost model performances (Clinical BERT alone: $SaO_2$ AUROC=0.9430; Clinical BERT with structured input: $SaO_2$ AUROC=0.9523). To our knowledge, the use of word position and layout in NLP models has not been reported. We proposed a new direction for optimizing NLP model performances.

Despite the promising findings, our study does have a few limitations. First, our data were secondarily collected. The cohort selection criterion emphasized patients with sleep disorders instead of all patients with a sleep study. However, as a methodology study that only focuses on scanned report processing, we believe patient demographics and medical history would not significantly affect our results or bias our methodological findings. Besides, due to the current insurance reimbursement environment, having a sleep disorder diagnosis is often a prerequisite for reimbursement of a sleep study. Hospitals often do not order sleep studies unless the patient is suspected of having sleep disorders. To some degree, our cohort captured the typical population who had sleep studies. Another limitation associated with the secondary data is the labels. The human chart review from the previous study was performed at the document level. The chart-reviewers did not collect the word-level annotation required for model training. To handle this, we matched AHI and $SaO_2$ values and label all exact-matched numbers in the text as targets. As a result, numbers that by chance had the same value as a target would be mislabeled as a target. This is possible as there are some other measurements (e.g., Apnea index, snore index) that have overlapping reference ranges, though this is still unlikely. Further, due to their importance in a sleep study report, the AHI and $SaO_2$ values could be mentioned multiple times. This made the challenge more complicated. This limitation would potentially hurt the model training by introducing noise. Fortunately, the document accuracy was evaluated solely with document-leveled labels and was not affected. We were able to show high performances even with this limitation.

Although AHI and $SaO_2$ values were often reported in writings, some reports presented them in tables, which is a 2-dimensional structure. Our text segmentation process only captures the sequence of words before and after the candidate numbers and does not incorporate words above or under them. Thus, column names in tables were not captured. Manually reviewing the target numbers that failed to be recognized by the NLP model, we found a high proportion of them were segmented from tables. Unfortunately, there was limited research about processing natural language in tables. In our data, tables are not often used for reporting AHI and $SaO_2$. Future studies are needed to resolve this problem.